\documentclass[10pt,twocolumn,letterpaper]{article}
\usepackage[dvipsnames]{xcolor}
\usepackage{iccv}
\makeatletter
\@namedef{ver@everyshi.sty}{}
\makeatother
\usepackage{tikz}
\usepackage{times}
\usepackage{epsfig}
\usepackage{graphicx}
\usepackage{amsmath}
\usepackage{amssymb}
\usepackage{etoolbox}
\usepackage{colortbl}
\usepackage{bbm}
\usepackage{pgfplots} 
\usepackage{pgfplotstable} 
\pgfplotsset{compat=newest} 
\usetikzlibrary{plotmarks} 
\usetikzlibrary{colorbrewer} 
\usepackage{pifont}
\usepackage{booktabs}

\usepackage[pagebackref=true,breaklinks=true,letterpaper=true,colorlinks,bookmarks=false]{hyperref}

\newcommand{\OURS}{CAD-Estate}
\newcommand{\LONGOURS}{\OURS: Large-scale CAD Model Annotation in RGB Videos}
\iccvfinalcopy 

\def\iccvPaperID{7535}

\ificcvfinal\fi

\begin{document}

\title{\LONGOURS}

\author{Kevis-Kokitsi Maninis\\Google Research \and Stefan Popov\\Google Research \and Matthias 
Nie{\ss}ner\\TUM 
\and Vittorio Ferrari\\Google Research
}

\maketitle
\ificcvfinal\thispagestyle{empty}\fi

\newcommand{\para}[1]{\par\noindent\textbf{#1}}

\newcommand{\showonerow}[6]{%
\setlength{\fboxsep}{0pt}
\resizebox{\textwidth}{!}{%
\fbox{\includegraphics[height=#6\linewidth]{figures/#5/#1.jpg}}
\hfill
\fbox{\includegraphics[height=#6\linewidth]{figures/#5/#2.jpg}}
\hfill
\fbox{\includegraphics[height=#6\linewidth]{figures/#5/#3.jpg}}
\hfill
\fbox{\includegraphics[height=#6\linewidth]{figures/#5/#4.jpg}}
}
}

\newcommand{\hlinesep}{\noindent\rule[7pt]{\linewidth}{0.4pt}\\}

\providetoggle{showcomments}
\settoggle{showcomments}{false} 								%


\iftoggle{showcomments}{
	\newcommand{\resolved}[3][]{\ifstrequal{#1}{resolved}{\textcolor{blue}{RESOLVED:}~\textbf{{\MakeUppercase
	 #2:}}~{#3}}{\textbf{\MakeUppercase #2:}~#3}}
	\newcommand{\stefan}[2][]{\textcolor{ForestGreen}{\resolved[#1]{stefan}{#2}}}
	\newcommand{\vitto}[2][]{\textcolor{red}{\resolved[#1]{vitto}{#2}}}
	\newcommand{\kevis}[2][]{\textcolor{orange}{\resolved[#1]{kevis}{#2}}}
	\newcommand{\MATTHIAS}[2][]{\textbf{\textcolor{red}{\resolved[#1]{MATTHIAS}{#2}}}}
	\newcommand{\changed}[1]{\textcolor{cyan}{#1}}
	\newcommand{\todo}[1]{\textcolor{blue}{\textbf{TODO:} #1}}
}{
	\newcommand{\changed}[1]{#1}
	\newcommand{\todo}[1]{}
	\newcommand{\stefan}[2][]{}
	\newcommand{\vitto}[2][]{}
	\newcommand{\MATTHIAS}[2][]{}
	\newcommand{\kevis}[2][]{}
}
\begin{abstract}
\vspace{-3mm}
We propose a method for annotating videos of complex multi-object scenes with a globally-consistent 3D representation of the objects. We annotate each object with a CAD model from a database, and place it in the 3D coordinate frame of the scene with a 9-DoF pose transformation. 
Our method is semi-automatic and works on commonly-available RGB videos, without requiring a depth sensor.
Many steps are performed automatically, and the tasks performed by humans are simple, well-specified, and require only limited reasoning in 3D. This makes them feasible for crowd-sourcing and has allowed us to construct a large-scale dataset by annotating real-estate videos from YouTube.
Our dataset \OURS{} offers \changed{101k} instances of 12k unique CAD models placed in the 3D 
representations of \changed{20k} videos.
In comparison to Scan2CAD, the largest existing dataset with CAD model annotations on real 
scenes, \OURS{} has \changed{7}$\times$ more instances and 4$\times$ more unique CAD models.
We showcase the benefits of pre-training a Mask2CAD model on \OURS{} for the task of automatic 
3D object reconstruction and pose estimation, demonstrating that it leads to performance improvements on the 
popular Scan2CAD benchmark.
\changed{
The dataset is available at {\footnotesize \url{https://github.com/google-research/cad-estate}}
}.
\end{abstract}

\section{Introduction}

Semantic 3D scene understanding from images and videos is a major topic in 3D scene understanding, crucial for many computer vision applications, ranging from robotics to AR/VR scenarios. The final goal is to detect all objects in the scene, recognize their class, reconstruct their 3D shape, as well as their pose within the overall scene coordinate frame.
With the advances of scalable deep learning techniques, the field has progressed from 
reconstructing the 3D shape of one object in a simple image with trivial 
background~\cite{mescheder19cvpr,wang18eccv,richter18cvpr,niu18cvpr,choy16eccv,girdhar16eccv},
to limited reasoning about object arrangements in simple multi-object 
scenes~\cite{popov20eccv,gkioxari19iccv,kuo20eccv},
and finally to unrestricted multi-object 3D reconstruction in complex real-world 
scenes~\cite{tyszkiewicz22eccv,maninis23tpami,runz20cvpr,rukhovich22wacv}.
This evolution has been dependent on the availability of ever larger and more diverse data sets for 
training and 
evaluation~\cite{avetisyan19cvpr,dai17cvpr,chang173dv,song15cvpr,sun18cvpr,shapenet15arxiv,li2018interiornet,fu21iccv,fu2021ijcv}

\begin{table*}[t]
\begin{minipage}{\linewidth}
\centering
\resizebox{1\linewidth}{!}{
\begin{tabular}{r|clcccc}
 Dataset                                    & Type of data & Sensor type   & Multi-object  & Annotation type   & Requires 3D reasoning & Total \# objects  \\ \hline
 SUN RGB-D~\cite{song15cvpr}                & image        & RGB-D         & \ding{51}     & 3D box            & yes                  & 64.6k                  \\
 PASCAL 3D+~\cite{xiang2014beyond}          & image        & RGB           & $\sim$        & CAD               & yes              & 36k                    \\
 IKEA~\cite{lim2013ikeaobjects}                    & image        & RGB           & \ding{55}     & CAD               & limited              & 759                    \\
 Pix3D~\cite{sun18cvpr}                     & image        & RGB           & \ding{55}     & CAD               & limited              & 10k                    \\
 ABO~\cite{collins2022abo}                  & image        & RGB           & \ding{55}     & CAD               & no                   & 6.3k                   \\ \hline
 Objectron~\cite{ahmadyan2021objectron}     & video        & RGB           & \ding{55}        & 3D box            & yes                  & 17k                    \\
 CO3D~\cite{reizenstein2021common}          & video        & RGB           & \ding{55}     & object point cloud      & no                   &19k                     \\
 Replica~\cite{straub2019replica}           & video        & RGB-D++       & \ding{51}     & labels on scene point cloud   & yes                  &$\sim$3k                \\
 Matterport3D~\cite{chang173dv}             & video        & RGB-D         & \ding{51}     & labels on scene point cloud   & yes                  &50.8k                   \\
 Scan2CAD~\cite{avetisyan19cvpr}            & video        & RGB-D         & \ding{51}     & CAD               & yes                  &14.2k                   \\
 \OURS{} (Ours)                             & video        & RGB           & \ding{51}     & CAD               & limited              & \changed{101k}                 \\
\end{tabular}
}
\vspace{0mm}
\caption{Real 3D scene understanding datasets and their attributes.
`Multi-object': whether there is more than one annotated object in the same image/video.
`Annotation type': what constitute the annotation for an object.
'Requires 3D reasoning': whether annotators need to reason in 3D.
'Total': number object instances with annotations.
}
\vspace{-2mm}
\label{tbl:all_datasets}
\end{minipage}
\end{table*}

Existing datasets for Semantic 3D scene understanding fall broadly in two categories: synthetic and acquired from real images/videos.
The former~\cite{li2018interiornet,fu21iccv,fu2021ijcv,song2017suncg,roberts21hypersim}
feature artificial 3D scenes that are manually designed by human artists, and then rendered into synthetic images.
While these datasets are relatively large, their images/videos expose a domain gap to real 
imagery~\cite{zakharov22eccv,richter23tpami,tremblay18TrainingDN,huang18eccv,richter16eccv,peng18syn2real}.

Acquired datasets
\cite{dai17cvpr,avetisyan19cvpr,song15cvpr,straub2019replica,chang173dv} annotate 3D objects on real images and videos (Table~\ref{tbl:general_statistics}).
Such datasets have been limited in size and diversity so far, partly due to limitations in their annotation process.
They rely on specialized equipment to capture depth images (RGB-D) in order to get a high-quality 3D point cloud reconstruction of the scene.
Humans then annotate objects on this 3D point cloud.
However, it is very expensive and cumbersome to go and physically acquire RGB-D videos in the real world, which limits the number of scenes captured, as well as their variety
(e.g. RGB-D sensors struggle outdoors due to sunlight, fail on glossy surfaces, and they have limited depth range).
Moreover, annotating on 3D point clouds requires expert annotators able to reason in 3D.

In this paper, we present the CAD-Estate dataset, which annotates real videos of complex scenes from Real Estate 10k \cite{zhou2018tog} with globally-consistent 3D representations of the objects within them. For each object we find a similar CAD model from a database, and place it in the 3D coordinate frame of the scene with a 9-DoF pose transformation.
We designed a semi-automatic approach which works on commonly-available RGB videos, without requiring a depth sensor, thereby opening the door to annotating many videos readily available on the web.
In our approach many steps are performed automatically, and the tasks performed by humans are simple, well-specified, and require only very limited reasoning in 3D. This makes them feasible for crowd sourcing, enabling to distribute work to a large pool of annotators. In turn, this has allowed us to construct a truly large-scale data set.

CAD-Estate contains \changed{100,882} instances of \changed{12,024} unique CAD models, covering \changed{19,512} videos (Sec.~\ref{sec:analysis}).
The models span 49 categories, 28 of which with more than 100 objects annotated.
In comparison, the largest existing dataset with CAD model annotations on real multi-object scenes (Scan2CAD~\cite{avetisyan19cvpr}) has
\changed{$7\times$} fewer objects (14,225),
$4\times$ fewer unique CAD models,
$2\times$ fewer categories with more than 100 objects (14) and
\changed{$13\times$} fewer videos (1,506).

In our experiments, we show that pre-training a modern model for automatic 3D object reconstruction and pose estimation~\cite{kuo20eccv} on \OURS{} improves performance on the popular Scan2CAD benchmark~\cite{avetisyan19cvpr}. Moreover, we establish baseline performance on our own test set, and provide ablation experiments to validate various choices of our annotation pipeline.

\section{Related Work}\label{sec:related_work}

\begin{figure*}[t]
  \centering
  \includegraphics[width=1\linewidth]{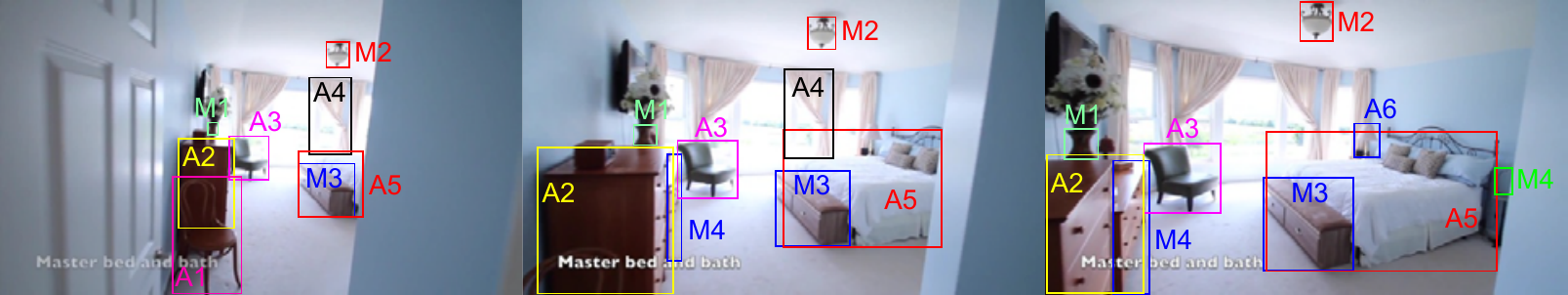} \\[0mm]
  \caption{{\small
  Automatic (A) and manual tracks (M). Automatic tracking can miss
  objects if they are truncated, occluded (like M4), or small (like M1).  We complete such
  tracks manually, on the validation and test sets.
  }}
  \vspace{-2mm}
  \label{fig:tracking}
\end{figure*}

\paragraph{Synthetic scene understanding datasets.}
Datasets of 3D object assets (without their poses on images) include ShapeNet~\cite{shapenet15arxiv}, 3D-FUTURE~\cite{fu2021ijcv}, ABC~\cite{koch2019abc} and ABO~\cite{collins2022abo}. Most recently, Objaverse~\cite{deitke2022objaverse} released a large dataset of 818k 3D assets.
Other synthetic datasets contain 3D objects placed in artificial 3D scenes designed by artists, usually indoor rooms~\cite{li2018interiornet,song2017suncg,roberts21hypersim}, and then rendered into images.

Synthetic datasets are large scale (up to 818k objects of~\cite{deitke2022objaverse}), but require extra efforts to bridge the domain gap for applications on real imagery~\cite{zakharov22eccv,richter23tpami,tremblay18TrainingDN,huang18eccv,richter16eccv,peng18syn2real}.

\paragraph{Real 3D scene understanding datasets.}
Several datasets have objects annotated on individual images (Table~\ref{tbl:all_datasets}, top block).
Sun RGB-D~\cite{song15cvpr} provides image-depth pairs from an RGB-D sensor along with objects annotated with 3D bounding-boxes (no 3D shapes).
PASCAL-3D+~\cite{xiang2014beyond} aligns simple CAD models to images by manually specifying the object pose and the focal length of the camera. They focus on simple images with fewer than 2 instances on average.
IKEA Objects~\cite{lim2013ikeaobjects} and Pix3D~\cite{sun18cvpr} annotated one object per image by aligning a 3D CAD model on it. Moreover, their scale is limited by the requirement for having CAD models exactly matching the objects in the images, which are difficult to find.
More recently, ABO~\cite{collins2022abo} automatically estimated 3D object poses for part of their 3D assets, on automatically retrieved images (6.3k images with one object annotated in each).

Other datasets annotate objects on videos (Table~\ref{tbl:all_datasets}, bottom block).
CO3D~\cite{reizenstein2021common} and Objectron~\cite{ahmadyan2021objectron} have videos mostly featuring one object each, and provide either a reconstructed point cloud of the object~\cite{reizenstein2021common} or a 3D bounding box~\cite{ahmadyan2021objectron}.
Several works~\cite{straub2019replica,chang173dv,dai17cvpr} use an RGB-D sensor to capture videos of rooms with multiple objects, then reconstruct a 3D point cloud scan of the scene by fusing the acquired depth maps.
They then label this 3D scan with object class and instance labels, resulting in incomplete object shapes.
Closer to our work, Scan2CAD~\cite{avetisyan19cvpr} goes a step further, building on~\cite{dai17cvpr} by manually annotating posed CAD models on the 3D scan.
These datasets heavily rely on a depth sensor, which limits their scale and applicability.
In contrast, we propose an annotation method which works on RGB videos, enabling annotating videos readily available on the web. Moreover, our human annotation tasks are very simple, and require little reasoning in 3D. These two features make our approach more scalable.
We construct \OURS{}, which annotates \changed{101k} objects with clean CAD models and full 9-DoF poses on pure RGB videos.
This is larger than any other dataset of real imagery, and is \changed{$7\times$} larger than Scan2CAD, which also offers posed CAD models (on RGB-D video).

\paragraph{Multi-object 3D reconstruction}
Many works tackle multi-object 3D reconstruction from a single image~\cite{gkioxari19iccv,popov20eccv,kuo20eccv,kuo2021patch2cad}. They are either trained on synthetic data~\cite{popov20eccv}, or on small real datasets~\cite{gkioxari19iccv, kuo20eccv, kuo2021patch2cad}.
Similarly, recent learning-based approaches reconstruct a scene from a video~\cite{maninis23tpami, li21odam, li2020ral, tyszkiewicz22eccv, runz20cvpr}, and use Scan2CAD as their main evaluation benchmark.
Our \OURS{} dataset can benefit all of these works as it offers new, large-scale, diverse, real video data with annotated complex spatial arrangements of 3D objects into scenes. In Section~\ref{sec:experiments} we show that pretraining on \OURS{} boosts the results of~\cite{kuo20eccv} on the original dataset it has been trained for~\cite{avetisyan19cvpr}.

\section{Dataset construction}
\label{sec:method}

Given a video of a static scene, our goal is to create a globally-consistent 3D representation that contains all its objects. To achieve this, we propose a semi-automatic system that relies on a large database of CAD models.
For each object in the scene, we find a similar-looking CAD model from the database and place it in the 3D coordinate frame of the scene by estimating its 9-DoF pose (i.e. 3D translation, 3D rotation, and 3D scale, allowing for independent scaling along each axis).

We design the system so that many steps are performed automatically. We leave only a few, simple and well-specified tasks for human annotators. These are all decomposed over individual objects, removing the complexities of considering the whole scene, and involve only very limited reasoning in 3D. These characteristics make the tasks feasible for crowd sourcing, enabling to distribute work to a large pool of annotators, as opposed to few in-house experts~\cite{dai17cvpr,avetisyan19cvpr}. This enables constructing a truly large dataset.

We annotated videos of RealEstate10K~\cite{zhou2018tog}, which show multiple rooms of real estate properties. The videos are split into shots, and camera poses have been extracted using an SfM pipeline~\cite{schonberger16cvpr}. We use ShapeNet~\cite{shapenet15arxiv} as our CAD model database, which contains 51k objects over 55 classes.

\paragraph{System overview.}
Our system receives an RGB video as input, with camera parameters for each frame (typically derived using SfM~\cite{schonberger16cvpr}). The output is the class, 3D pose (rotation, translation, scale), and 3D shape of each object in the video (represented as a CAD model from a database).

The system amounts to a sequence of 5 stages:
\\
(1) We start by detecting objects in the video and tracking them over time, either automatically or with the help of humans (Sec.~\ref{sec:method-tracking}). Each track corresponds to one physical object in the scene and forms the unit of annotation. All further stages operate on one track at a time with the goal of reconstructing its pose, shape, and class.
\\
(2) For each track, we automatically select a few similar-looking CAD model candidates from the database, and then ask humans to choose the best match (Sec.~\ref{sec:method-cad-select}).
\\
(3) We ask humans to annotate $3D \leftrightarrow 2D$ point correspondences between the chosen CAD model and the object in the video, on a few key-frames (Sec.~\ref{sec:method-pca}).
\\
(4) We use the annotated correspondences together with the camera parameters of the key-frames to automatically estimate the 9-DOF pose of the object (Sec.~\ref{sec:method-optimization}).
\\
(5) Finally, we ask humans to verify the estimated pose for quality control (Sec.~\ref{sec:method-verification}).

\subsection{2D Object detection and tracking}
\label{sec:method-tracking}

In this first stage we detect objects in the video and track them over time. Each track then corresponds to one physical object and forms the unit of annotation for all subsequent stages.
We apply somewhat different procedures for the training and val/test sets of our dataset, in order to strike a good trade-off between automation (hence reducing human effort) and completeness of annotation (we want to capture all objects in the val/test set).

\paragraph{Train set.}
We detect objects in each frame automatically using a SpineNet-based model~\cite{du20cvpr}.

We also extract an appearance descriptor for each detection box, by applying a Graph-Rise-based~\cite{juan19arxiv} model.

Next, we associate detections over time, as common in tracking-by-detection 
approaches~\cite{bergmann2019tracking,danelljan2019atom,andriluka2008people}.
We compute various similarity scores between two detections in different video frames, including the similarity between their appearance descriptors, the difference in their class labels, and the spatial continuity of the box positions in adjacent frames.
Then we cluster all detections across all frames into tracks based on these similarity scores using the Clique Partitioning approach of \cite{marin14ijcv}.

\paragraph{Val/Test sets.}
Automatic detection and tracking models can sometimes miss objects as they do not work perfectly.
Since for the validation and the test sets we strive for a high degree of completeness, we annotate missing object tracks manually (in addition to the automatic ones).
For this we developed an efficient custom interface that allows annotators to draw a whole object track in time, i.e. drawing a bounding-box~\cite{papadopoulos17iccv} on each key-frame where a particular physical object appears. For efficiency, we automatically focus work on $6$ frames regularly-spaced in time.
The annotators see all current tracks already found by the automatic approach, and only draw 
missing ones.

Note how we apply this manual annotation procedure only to a rather small subset of the data (val/test sets have fewer videos than train, Table~\ref{tbl:general_statistics}).

\subsection{Selecting a CAD model}
\label{sec:method-cad-select}

The second stage is to select a suitable CAD model for a tracked object. We first select $10$
candidates automatically from the database.
We then ask a human to chose the one that looks the closest to the object in the video. This removes the need for annotators to search through the large database.

\paragraph{Finding candidates automatically.}
We find candidate CAD models for an object track by considering both appearance similarity and class label similarity cues.
During pre-processing, we render the CAD models in the database from 10 random viewpoints and compute an appearance descriptor for each view (the same as in Sec. \ref{sec:method-tracking}). We then compute the appearance similarity between an object box in a frame of the object track and a CAD model view as the cosine similarity of their descriptors.

For the class label similarity we need to take special care, as the label spaces of the CAD model database and the object detector are different and feature multi-way relationships
(e.g. the CAD "cabinet" matches the detector's "filing cabinet", "wardrobe", and "chest of drawers").
Hence, we embed each class label name into a common semantic space using the Universal 
Sentence Encoder~\cite{cer18emnlp}, and compute the cosine similarity between any two class 
labels in this space. This is a general solution that can work with any label space.
We combine the appearance and class similarity scores with a simple product.

To compute the overall similarity between an object track and a CAD model, we aggregate the combined appearance-class similarity over all pairs of frames and CAD model views.
We use this overall similarity score to rank CAD models and select the top 10 as candidates for an object track.
In practice the class similarity act as a soft filter for the appearance similarity, so the best CAD models are the most similar-looking ones to the object in the track, among those that have a similar class label.

\paragraph{Selecting the best candidate with a human.}
We ask annotators to choose the best matching candidate. We show them the detected object on a set of evenly spaced key-frames, next to the rendered CAD model candidates. Annotators can navigate between key-frames, to see the object from multiple views. Annotators can declare that none of the candidates are similar enough to the tracked object (hence that track is not passed on to the later stages).

\subsection{3D $\leftrightarrow$ 2D point correspondences}
\label{sec:method-pca}

\begin{figure}
\centering
\includegraphics[width=1\linewidth]{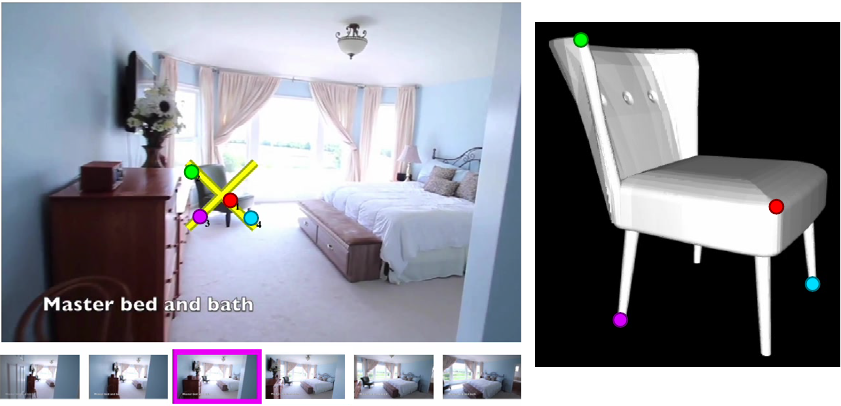} \\[1mm]
\caption{{\small
We ask humans to annotate point correspondences between the 3D surface of the CAD model and the video
frames of the tracked object.
}}
\vspace{-0mm}
\label{fig:ui-pca}
\end{figure}

We now ask humans to annotate point correspondences between the 3D surface of the CAD model and the video frames of the tracked object (Fig.~\ref{fig:ui-pca}).
As for the CAD candidate selection case, the interface enables annotators to navigate between key-frames. We show the selected CAD model next to the key-frames.
For each key-frame, we ask annotators to annotate 4-6 point correspondences between the CAD model and the frame. To make the task easier, they can rotate and flip the CAD model in 3D, in order to roughly match the orientation of the object in the frame.
We will use these correspondences to recover the 9-DOF object pose in the next stage.

Our approach consists of steps that are easy to understand and easy to master.
Annotators control rotation with Orbit Controls~\cite{livingston00jgt}, which translates 2D mouse 
movements to view-local object rotation in 3D in an intuitive way.
Afterwards, clicking on CAD-to-image point correspondences is very easy and is similar to other familiar 2D annotation tasks.
Most importantly, this approach is object-centric and requires no reasoning in 3D in the global coordinate frame of the scene. Instead, this harder task is done automatically in the pose estimation stage of our system.
Finally, annotating point correspondences is decoupled between frames: the annotator is free to pick different points in every frame. This makes it easy even for objects with complex shapes.

\subsection{Object 3D pose estimation}
\label{sec:method-optimization}

We use the $3D \leftrightarrow 2D$ point correspondences to automatically estimate a global 9-DOF pose for the object. We apply a non-linear optimization method, which integrates evidence from all views in a track, and consists of multiple objectives.

We express the object pose as a 9-DOF transformation that brings the CAD model from its canonical pose to the world coordinate frame of the scene.

The transformation has 3 components: 3D translation $T$, 3D rotation $R$, and anisotropic 3D scale $S$ (i.e. we allow independent scaling along each axis).
The goal is to recover this unknown transformation $(T,R,S)$. We setup below several objectives, which are functions of $(T,R,S)$, and combine them into an overall objective. Finally we minimize that overall objective over $(T,R,S)$.

\begin{figure}
\centering
\includegraphics[width=1\linewidth]{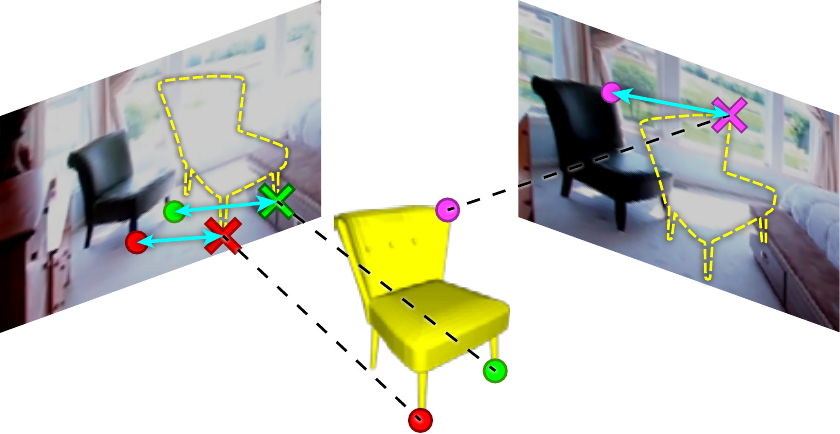} \\[1mm]
\vspace{-0mm}
\caption{{\small
Point re-projection objective. We project the annotated points on the CAD model to the video frames given a candidate pose $(T,R,S)$, and penalize the displacement with respect to their 2D correspondences (arrows in cyan). We minimize this objective over poses (along with two others). 
}}
\vspace{-0mm}
\label{fig:multiview_integration}
\end{figure}

\paragraph{Point re-projection objective.}
We know the extrinsic and intrinsic camera parameters at each video frame.
Given a potential $(T,R,S)$ we can use it along with the camera parameters to project the 3D points on the surface of the CAD model to the video frame. Therefore, we setup a point re-projection objective $L_{repr}(T,R,S)$ which measures the L1 distance between the projected 3D points and their corresponding 2D points in each frame (and sum over all frames, Figure~\ref{fig:multiview_integration}).
The correspondences are given by the $3D \leftrightarrow 2D$ annotations from Sec. \ref{sec:method-pca}, and we also take into account whether the annotator flipped the CAD model.

\paragraph{Up-axis objective.}
Most objects in our videos are usually placed vertically in an upright position. We reflect this by imposing an $L1$ objective that penalizes 3D rotations that change the "up"-axis of the object with respect to the world.
We do this directly on the target rotation matrix $R$ by applying the additional objective $L_{up}(R)$.
This objective is applied to object classes that are usually found in upright position (e.g. chairs, tables, cabinets, etc.), whereas other classes such as pillows are excluded.
For this objective to be applied, we need to know the up-axis for the objects in our CAD database, 
and in the world coordinate frame
(which we do for ShapeNet and RealEstate10K).

\paragraph{Front-of-camera objective.}
We encourage object pose transformations that place all annotated 3D points in front of their respective cameras (rather than behind), by penalizing 3D points that have a negative depth in the coordinate frame of that camera.

\paragraph{Special scale parameterization for co-planar 3D points.}
Sometimes, all 3D points chosen in Sec. \ref{sec:method-pca} by the annotator on the CAD model are co-planar. This typically happens when the video shows only a planar part of the object, e.g. a table seen only from the top, or a cupboard seen only frontally. Co-planar 3D points prevent resolving all three dimensions of the target scaling transformation $S$.
We detect such cases automatically during annotation. We then resolve them during pose estimation by constraining the scaling factor perpendicular to the annotated plane to be the average of the other two scale factors. This reduces the DOF of the scaling transformation $S$ down to 2.

\paragraph{Special rotation/scale parameterization for symmetric objects.}
In many cases the retrieved CAD models are symmetric, which typically leads to inconsistent point correspondence annotations across frames (e.g. an annotator picking a particular 3D point on a rotation-symmetric lamp corresponds to a point in the video in a frame, but then picking a different 3D point in a different frame, as these are equivalent up to symmetry).
We handle these cases by optimizing for a rotation w.r.t any of the symmetries of the object in the reprojection objective.
We consider the same symmetries as in Scan2CAD, i.e. 2-way (e.g. a rectangular table), 4-way (e.g. a square table), and 36-way (e.g. a round table).
We detect symmetries automatically directly on each CAD model.
For fully symmetric objects (36-way symmetric), we further constrain the two scaling factors around the up-axis to be identical.

\paragraph{Optimization}
We combine the above objectives in an overall one:
\begin{equation}
\small
\begin{split}
L_{pose}(T,R,S) = \ & L_{repr}(T,R,S) + \alpha \cdot L_{up}(R) \\
                & +\beta \cdot L_{front}(T,R,S)
\end{split}
\end{equation}

We minimize this objective over $(T,R,S)$ with Adam~\cite{kingma2014adam}. $\alpha$ and $\beta$ are hyperparameters set empirically.

\subsection{Pose verification by humans}
\label{sec:method-verification}

In this last stage, we verify whether the pose computed in the previous stage matches the image contents in the video. This is necessary as pose estimation can fail for several reasons, including limited/degenerate camera motion, occlusion, and truncated objects.

We render the CAD model as overlay on top of the video frames in a track, using the camera 
parameters and the estimated object pose $(T,R,S)$. We then ask human annotators to judge 
whether the rendered CAD aligns well with the object in the video. If it aligns well in all key-frames, 
we mark the pose as correct.

\begin{figure}
\centering
\pgfplotstableread{data/truncation_histogram.txt}\truncdata
\mbox{%
\begin{minipage}{1\linewidth}
  \resizebox{\textwidth}{!}{%
    \begin{tikzpicture}
        \begin{axis}[set layers,width=1.4\linewidth,height=0.45\linewidth,
                ybar=0pt,bar width=0.3,
                grid=both, grid style=dotted,
                ytick={0,.2,.4,.6},
                scaled y ticks=false,
                ymin=0, ymax=0.65,
                xtick=data,
                xticklabels={0\%,10\%,20\%,30\%,40\%,50\%,60\%,70\%,80\%,90\%},
                x tick label style={rotate=0,anchor=north east,xshift=0pt,yshift=5pt},
                major x tick style = transparent,
                enlarge x limits=0.09,
                font=\normalsize,
                ]

            \addplot[draw opacity=0,fill=Set2-8-1,mark=none,legend image post style={yshift=-0.1em}] table[x expr=\coordindex,y=CadEstate]{\truncdata};
            \label{fig:trunc:cadestate}

            \addplot[draw opacity=0,fill=Set2-8-4,mark=none,legend image post style={yshift=-0.1em}] table[x expr=\coordindex,y=Scan2CAD]{\truncdata};
            \label{fig:trunc:scan2cad}
            \legend{\OURS{}, Scan2CAD~\cite{avetisyan19cvpr}}
        \end{axis}
    \end{tikzpicture}
    }
    \end{minipage}

    }
    \vspace{2mm}
    \caption{\small{Truncation histogram for \OURS{} and Scan2CAD.} We measure the degree of object truncation as the fraction of a CAD model surface that projects outside of the video frame.}
    \vspace{-3mm}
    \label{fig:trunc}
\end{figure}
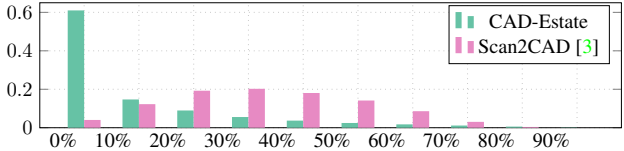
\section{Dataset analysis}
\label{sec:analysis}

\begin{figure*}
\pgfplotstableread{data/class_histogram.txt}\perseqdata
\mbox{%
\begin{minipage}{1.\textwidth}
  \resizebox{\textwidth}{!}{%
    \begin{tikzpicture}
        \begin{semilogyaxis}[set layers,width=1.4\textwidth,height=0.3\textwidth,
                ybar=0pt,bar width=0.4,
                grid=both, grid style=dotted,
                log origin=infty,
                minor ytick={1,2,...,10,20,30,...,100, 200,...,1000, 2000,...,10000, 20000},
                ytick={1, 10, 100,1000,10000, 25000},
                yticklabels={1, 10,100,1000,10000,25000},
                scaled y ticks=false,
                ymin=.7, ymax=30000,
                xtick = data, x tick label style={rotate=90,anchor=north east,xshift=4pt,yshift=8pt},
                xticklabels from table={\perseqdata}{class},
                major x tick style = transparent,
                enlarge x limits=0.02,
                font=\normalsize,
                ]

            \addplot[draw opacity=0,fill=Set2-8-1,mark=none,legend image post style={yshift=-0.1em}] table[x expr=\coordindex,y=cadestate]{\perseqdata};
            \label{fig:perseq:cadestate}

            \addplot[draw opacity=0,fill=Set2-8-4,mark=none,legend image post style={yshift=-0.1em}] table[x expr=\coordindex,y=scan2cad]{\perseqdata};
            \label{fig:perseq:scan2cad}
            \legend{\OURS{}, Scan2CAD~\cite{avetisyan19cvpr}}
        \end{semilogyaxis}
    \end{tikzpicture}
    }
    \end{minipage}
    }
    \vspace{0mm}
    \caption{Class histogram of \OURS{} vs. Scan2CAD~\cite{avetisyan19cvpr}. We annotate more classes with many more objects. Note the logarithmic scale of the vertical axis.}
    \vspace{2mm}
    \label{fig:class_histogram}
\end{figure*}

\paragraph{General statistics.}
Table~\ref{tbl:general_statistics} compares general statistics of \OURS{} to the closest existing 
video dataset Scan2CAD~\cite{avetisyan19cvpr}.
We further split the stats of our dataset into training set and val/test test sets.

\OURS{} is an order of magnitude larger than Scan2CAD (\changed{20k} vs. 1.5k scenes, and \changed{101k} vs. 14.2k posed objects). The annotated objects cover more classes (49 vs. 35 in Scan2CAD).
Figure~\ref{fig:class_histogram} shows the distribution of annotated objects over classes.
Despite the long tail, there are many more classes that have a large number of objects (13 classes with $>1000$ objects vs 4 in Scan2CAD, and 28 classes with $>100$ objects vs 14).

\OURS{} also offers greater diversity of object 3D shapes. It is annotated with \changed{12k} CAD models vs 3k for Scan2CAD (noting that in both datasets the CAD shapes are a close match rather than exactly matching the shape of the object in the image).

\begin{table}[t]
\begin{minipage}{\linewidth}
\centering
\resizebox{\linewidth}{!}{
\begin{tabular}{r|cc>{\columncolor[gray]{0.9}}c|c}
& \multicolumn{3}{c|}{\OURS} &  \\
Dataset               & Train       & Val/Test           & Total            & Scan2CAD   \\ \hline
\#Scenes              & \changed{16713} & \changed{2799} & \changed{19512}            & 1506       \\
\#Posed objects       & \changed{77832} & \changed{23050}& \changed{100882} & 14225      \\
\#Classes             & 49              & 40             & 49               & 35         \\
\#Classes $>1000$     & \changed{11}    & 5              & 13               & 4          \\
\#Classes $>100$      & 24              & 22             & 28               & 14         \\
\#Objects per scene   & 4.7             & 8.2            & 5.2              & 9.4        \\
\#CAD models          & \changed{10358} & \changed{6192} & \changed{12024}  & 3049       \\
\#Frames per scene    & \changed{138}   & \changed{139}  & \changed{138}    & 1604       \\
Source                & RGB             & RGB            & RGB              & RGB-D      
\end{tabular}
}
\vspace{0mm}
\caption{General statistics of \OURS{} and Scan2CAD.}
\vspace{-0mm}
\label{tbl:general_statistics}
\end{minipage}
\end{table}

\paragraph{Camera framing.}
There is a qualitative difference between the video captures of Scan2CAD (from ScanNet~\cite{dai17cvpr}) and \OURS{} (from RealEstate10K~\cite{zhou2018tog}).
The videos of~\cite{dai17cvpr,avetisyan19cvpr} were captured with an RGB-D sensor, taking close-up views which facilitates acquiring good quality depth maps.
Instead, the videos of \OURS{} are captures of real estate properties with more distant views that depict a larger part of each room, as the goal was to showcase the space for selling it.
The video shots are also shorter (\changed{138} frames per video in \OURS{} vs. 1.6k in Scan2CAD).

As a consequence of the more distant views, several key statistics are different in \OURS{}, compared to Scan2CAD:
(1) More objects are visible in one video frame at the same time: on average, 7.9 in \OURS{} vs 3.3 in Scan2CAD.
(2) More objects are further way from the camera and thus appear smaller on the images: on average, the bounding-box of a CAD-Estate object covers 7.5\% of the 
image area vs. 16.5\% in Scan2CAD.
(3) The dynamic range of the Z position of objects is larger: in CAD-Estate the farthest object is $4.5\times$ farther from the camera than the nearest  one, vs. $2.3\times$ in Scan2CAD.
(4) Object truncation is much higher in ScanNet compared to \OURS{}, where most objects are completely visible (Figure~\ref{fig:trunc}).
This is also a consequence of the capture process, as ScanNet needs close-up captures due to the range of the depth sensor.

The camera framing statistics above highlight how \OURS{} poses a different challenge than 
Scan2CAD for automatic scene understanding methods, as they need to handle more complex views with more objects visible at the same time, many smaller objects, a higher variability of their distance to the camera, but also less truncated by the image frame.

\section{Experiments}
\label{sec:experiments}

\begin{figure*}
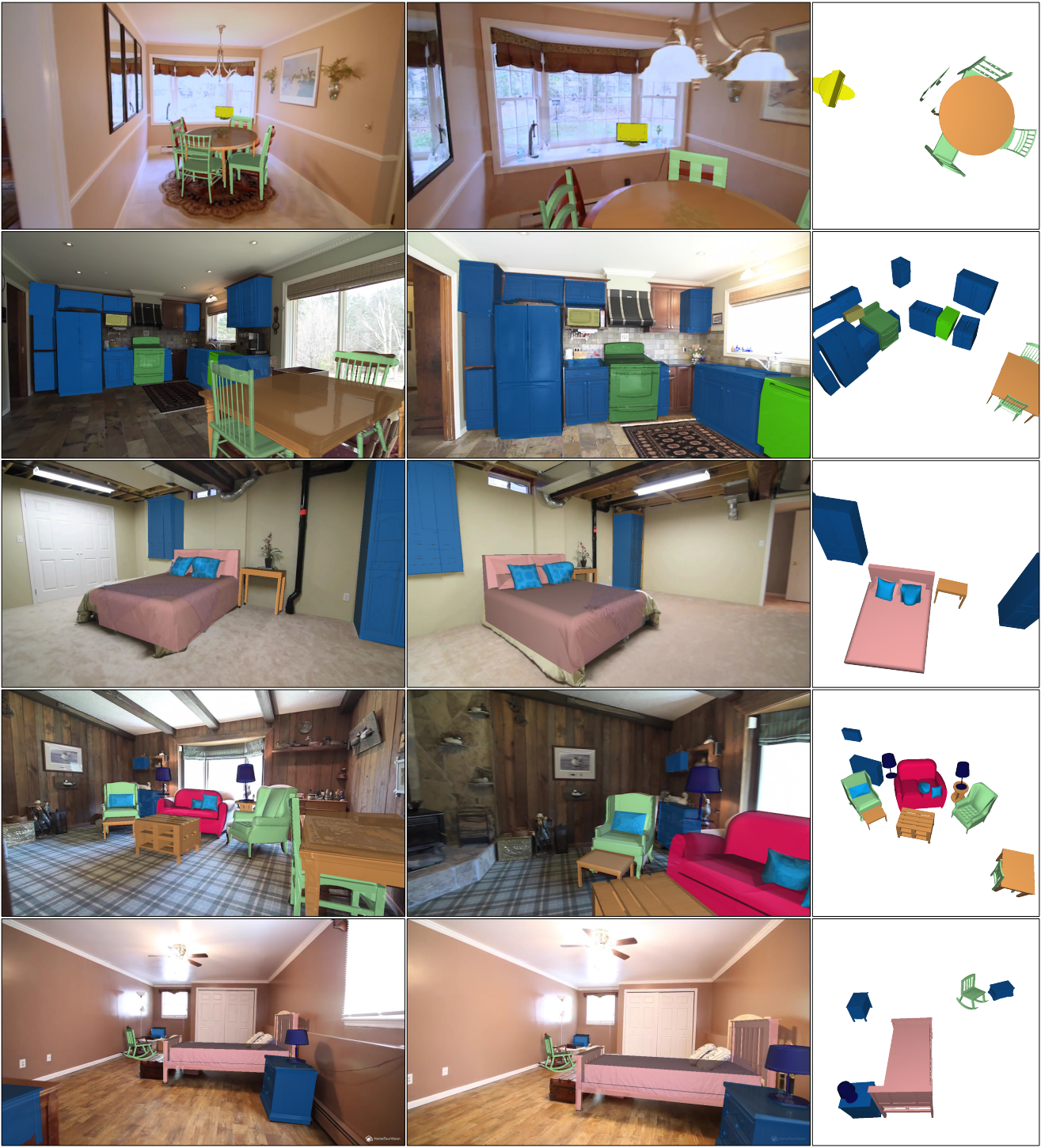

	\centering
		\showonecadestate{5LToC4KInNM_84017000}{0}{5}{3}{0.6}
		\showonecadestate{5RD3EAlBS9w_180880000}{1}{5}{2}{0.6}
		\showonecadestate{video_-AplGqzOF5Y_147046900}{0}{6}{2}{0.6}
		\showonecadestate{5RD3EAlBS9w_112012000}{2}{7}{2}{0.6}
		\showonecadestate{video_-4czxXKNrnI_199665000}{0}{6}{0}{0.6}
		\label{fig:qual:cadestate_qualitative_2}
	\caption{Annotated scenes from \OURS{}, overlaid on video frames (left, mid), and shown from a top view (right).}
	\vspace{20mm}
\end{figure*}

\begin{table*}
\begin{minipage}{\linewidth}
\centering
\resizebox{1\linewidth}{!}{
\begin{tabular}{c|>{\columncolor[gray]{0.9}}ccc|cccccccc}
Pretraining on \OURS & $AP_{mesh}$& $AP_{mesh}^{50}$& $AP_{mesh}^{75}$& bed &	sofa &chair &cabinet &bin &display &table &bookshelf  \\\hline
no (original~\cite{kuo20eccv}) & 8.4 & 23.1& 4.9&  14.2 &13 &13.2 &7.5 &7.8 &5.9 &2.9 &3.1 \\
no                         & 8.2 & 23.1& 4.9&  14.0 &12.7 &12.9 &7.1 &7.6 &6 &2.5 &3.0 \\
yes                        & \textbf{9.4} & \textbf{25.0} & \textbf{5.7}& 15.1 &13.2 &14.5 &9.0 &7.4 &7.8 &4.0 &4.5 \\
\end{tabular}
}
\vspace{3mm}
\caption{Performance of Mask2CAD on Scan2CAD's test set.
Top row:    results reported by~\cite{kuo20eccv} by training on Scan2CAD train set;
Second row: our reproduction of that experiment, which reaches nearly identical performance.
Bottom row: pre-training on \OURS{} train set, then fine-tuning on Scan2CAD train set. Performance improves thanks to our additional training data.
}
\vspace{-0mm}
\label{tbl:mask2cad_scan2cad}
\end{minipage}
\end{table*}

We first perform several experiments by training a learning-based method for CAD model alignment~\cite{kuo20eccv} on \OURS{} (Sec. \ref{sec:mask2cad-models}), demonstrating that it leads to performance improvements on the Scan2CAD test set, and establishing that our test set offers a harder challenge.
Then in Sec. \ref{sec:ablations} we provide ablation experiments for the different components of our 
annotation pipeline, showing their relative merit and demonstrating that they are all necessary to 
achieve high quality.

\subsection{Training Mask2CAD on \OURS{}}
\label{sec:mask2cad-models}

In this section, we showcase how \OURS{} can be used to train Mask2CAD~\cite{kuo20eccv}, a deep learning method for single-image 3D object reconstruction and pose estimation.
We start by studying the benefits of having a large training set by pre-training~\cite{kuo20eccv} on \OURS{} and then fine-tuning and evaluating on Scan2CAD~\cite{avetisyan19cvpr} (where Mask2CAD was originally benchmarked on).
Then we establish baseline results for Mask2CAD trained and tested on \OURS{}.
 
\paragraph{From \OURS{} to Scan2CAD.}
Mask2CAD has been extensively evaluated \cite{kuo20eccv} by training and testing on the Scan2CAD dataset~\cite{avetisyan19cvpr}, whose training set consists of 9.5k objects over 19k frames on 1194 scenes.
We run the same experiment, but first pre-train Mask2CAD on a much larger training set of 45k objects over 150k frames sampled from 11k scenes of \OURS's trainval. Then we fine-tune on the train set of Scan2CAD, and evaluate on the test set with the popular metrics $AP_{mesh}$, $AP_{mesh}^{50}$, and $AP_{mesh}^{75}$~\cite{kuo20eccv, gkioxari19iccv}.

Table~\ref{tbl:mask2cad_scan2cad} presents the results on all 3 metrics above, and additionally per-class $AP_{mesh}$.
As the results show, pre-training on our large dataset improve the performance of Mask2CAD significantly, for almost all classes. We observe that the improvement is greater for classes for which \OURS{} has many objects (cabinet, table, bed).

\begin{table*}
	\begin{minipage}{\linewidth}
		\centering
		\resizebox{1\linewidth}{!}{
			\begin{tabular}{c|>{\columncolor[gray]{0.9}}ccc|cccccccc}
				& $AP_{mesh}$& $AP_{mesh}^{50}$& $AP_{mesh}^{75}$& bed &	sofa &chair &cabinet 
				&bin &display &table &bookshelf  \\\hline
				Maks2CAD on \OURS{} &7.5 &21.2 &2.4 &13.4 &10.2 &10.7 &10.3 &2.0 &4.1 &5.2 &4.2 \\
			\end{tabular}
		}
		\vspace{2mm}
		\caption{Mask2CAD results on \OURS{}.}
		\label{tbl:mask2cad_re10k}
	\end{minipage}
\end{table*}

\paragraph{Train and test on \OURS{}.}
We now establish baseline results for Mask2CAD on \OURS{} (training on our trainval set, and evaluating on the test set).
We use the same classes as Scan2CAD for this experiments, and the same evaluation metrics, enabling approximate comparisons across datasets.

The results in Table~\ref{tbl:mask2cad_re10k} show that Mask2CAD achieves considerably lower performance on \OURS{} than on Scan2CAD.  Especially on the strict IoU threshold $AP_{mesh}^{75}$, the performance is much lower (5.7 vs. 2.4).
This indicates that our test set might offer a harder challenge. \OURS{} provides more complex 
scenes that are difficult to reconstruct, and objects are in general further away from the camera, 
which makes pose estimation harder.

\subsection{Optimization objectives for 3D pose estimation}
\label{sec:ablations}

We study the influence of the object pose optimization objectives of 
Section~\ref{sec:method-optimization} on pose estimation quality. We evaluate by asking 
annotators to verify the poses produces by different versions of the pose estimator (as in Sec. 
\ref{sec:method-verification}, but on a subset of the data). A higher percentage of positively verified 
object poses indicates a better pose estimator.

Starting from 52.2\%, the percentage of positively verified poses improves steadily as we add the special parameterization of the re-projection objective for handling co-planar 3D points (57.6\%), the one for handling symmetric objects (62.9\%), and the up-axis objective (74.9\%). This demonstrates that all of them contribute to the quality of our dataset, as they enable to estimate a correct pose for a greater number of objects.
The largest contribution is made by the up-axis objective, as it affects many objects. Instead, $27.8\%$ of all objects in our dataset are symmetric, and only $15.5\%$ received co-planar 3D point annotations.

\section{Conclusions}
\label{sec:conclusions}

We introduced a new way to annotate 9-DoF pose of CAD models on monocular RGB videos.
As a result of our method, we obtained the \OURS{} dataset, which features \changed{101k} instances of
12k unique CAD models placed in the 3D representations of \changed{20k} videos.
This dataset is an order of magnitude larger than existing CAD annotation efforts facilitated by our 
new annotation method.
We have shown experimentally that the quantity and diversity of such data significantly benefits the 
modern CAD alignment technique Mask2CAD, leading to improved performance on Scan2CAD.
However, we believe that this is only a first step, and \OURS{} is an important stepping stone 
towards leveraging CAD priors for 3D scene reconstruction and understanding in the context of a 
wide range of downstream tasks.

\paragraph{Acknowledgements:}
We thank Prabhanshu Tiwari, Sweety Chaudhary, Abha Dwivedi, Ashlesha Shantikumar, Umesh 
Vashisht, Mohd Adil for coordinating the annotation process, and Weicheng Kuo who helped us with 
running Mask2CAD on our dataset.

{\small
\bibliographystyle{ieee_fullname}
\bibliography{shortstrings,loco,loco_extra}
}

\end{document}